\pdfoutput=1

\documentclass[11pt]{article}
\usepackage[final]{acl}
\usepackage{tabularx}
\usepackage[utf8]{inputenc}
\usepackage[arabic,english]{babel}  
\usepackage{textcomp}
\usepackage{longtable}
\usepackage{subcaption}
\usepackage{graphicx}
\usepackage{placeins} 
\usepackage{float}
\usepackage{times}
\usepackage{latexsym}
\usepackage{booktabs}
\usepackage[T1]{fontenc}

\usepackage{multirow}

\usepackage{colortbl}
\usepackage{arydshln}
\usepackage{nicematrix}
\usepackage{microtype}
\usepackage{inconsolata}

\usepackage{hyperref}

\title{
\raisebox{-2.1ex}{\protect\includegraphics[height=4.5\fontcharht\font`\B]{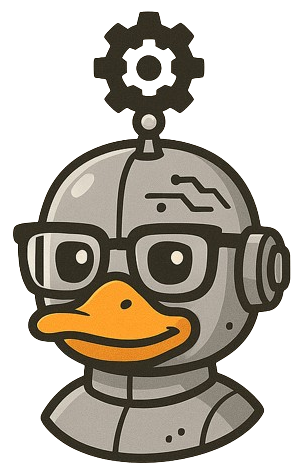}} AraReasoner: Evaluating Reasoning-Based LLMs for Arabic NLP
}

\definecolor{dkgreen}{rgb}{0,0.6,0}

\author{
  \textbf{Ahmed Hasanaath\textsuperscript{1}},
  \textbf{Aisha Alansari\textsuperscript{1}},
  \textbf{Ahmed Ashraf\textsuperscript{1}},
  \textbf{Salmane Chafik\textsuperscript{3}},\\
  \textbf{Hamzah Luqman\textsuperscript{1,2}},
  \textbf{Saad Ezzini\textsuperscript{1}}\\
  \textsuperscript{1}King Fahd University of Petroleum and Minerals\\
  \textsuperscript{2}SDAIA--KFUPM Joint Research Center for Artificial Intelligence\\
  \textsuperscript{3}Mohammed VI Polytechnic University\\
  \small\textbf{Correspondence:} \href{mailto:hluqman@kfupm.edu.sa}{\texttt{hluqman@kfupm.edu.sa}} 
}

\begin{document}
\maketitle
\vspace*{0.3em}
\section*{\centering ABSTRACT}

Large language models (LLMs) have shown remarkable progress in reasoning abilities and general natural language processing (NLP) tasks, yet their performance on Arabic data, characterized by rich morphology, diverse dialects, and complex script, remains underexplored. This paper presents a comprehensive benchmarking study of multiple reasoning-focused LLMs, with a special emphasis on the newly introduced DeepSeek models, across a suite of fifteen Arabic NLP tasks. We experiment with various strategies, including zero-shot, few-shot, and fine-tuning. This allows us to systematically evaluate performance on datasets covering a range of applications to examine their capacity for linguistic reasoning under different levels of complexity. 
Our experiments reveal several key findings. First, carefully selecting just three in-context examples delivers an average uplift of over 13 F1 points on classification tasks—boosting sentiment analysis from 35.3 \% to 87.5 \% and paraphrase detection from 56.1 \% to 87.0 \%. Second, reasoning-focused DeepSeek architectures outperform a strong GPT o4-mini baseline by an average of 12 F1 points on complex inference tasks in the zero-shot setting. Third, LoRA-based fine-tuning yields up to an additional 8 points in F1 and BLEU compared to equivalent increases in model scale. 
The code is available at \href{https://github.com/gufranSabri/deepseek-evals}{Project Repository} 



\section{Introduction}
\FloatBarrier  
Arabic is spoken by more than 400 million people across 22 countries, making it one of the most widely used languages in the world. Despite its global significance, Arabic has traditionally been underrepresented in natural language processing (NLP) research \cite{alturayeif2022mawqif}. This gap is partially due to the inherent complexity of the language, which features a rich morphological system, multiple dialects, and a non-Latin script. For instance, modern standard Arabic (MSA) differs considerably from various colloquial dialects in syntax and vocabulary, increasing the difficulty for models to generalize across tasks such as sentiment analysis, summarization, and translation. Moreover, diacritization and transliteration tasks pose unique challenges, as they require a nuanced understanding of phonological rules and script variations. These factors underscore the importance of dedicated research efforts to improve Arabic NLP capabilities. 

\begin{figure}[t]
    \centering
    \includegraphics [width=0.9\columnwidth]{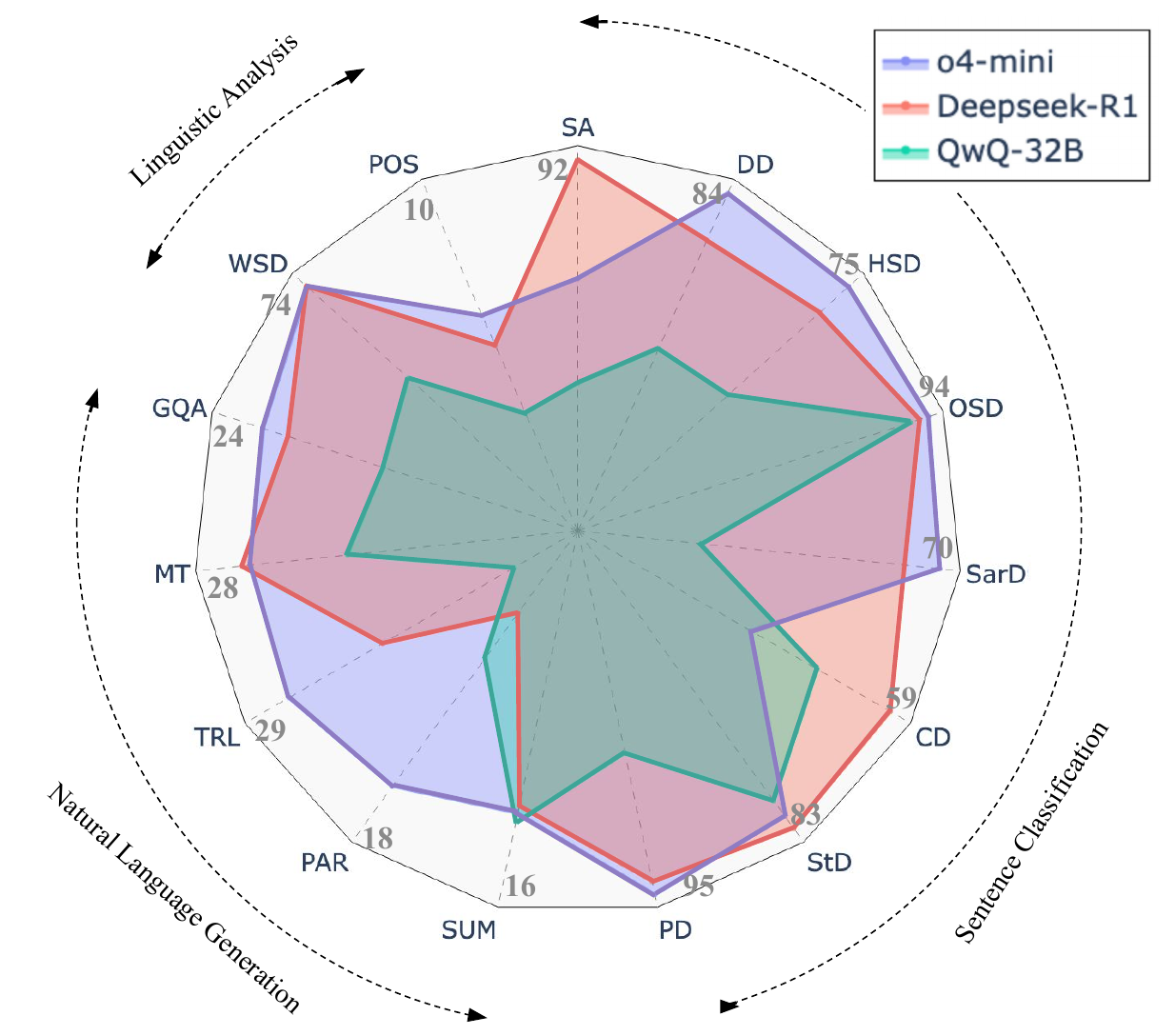}
    \caption{Zero-shot performance of reasoning-based models on 15 Arabic tasks, grouped into Sentence Classification, Linguistic Analysis, and Natural Language Generation. 
    }
    \label{fig:promptTemplate}
\end{figure}

Large-scale language models (LLMs) have demonstrated outstanding performance across a wide range of NLP tasks \cite{chang2024survey}. Models from the GPT \cite{radford2018gpt}, Gemini \cite{team2023gemini}, Deepseek \cite{bi2024deepseek}, and LLaMA \cite{touvron2023llama} families have shown exceptional capabilities in different NLP tasks. These models exhibit a strong understanding of natural language and express fluently in human language \cite{alyafeai2306taqyim}. This performance can be attributed mainly to their model architecture and the pre-training on large-scale corpora.
Recently, the newly introduced reasoning-based models, such as DeepSeek R1 and GPT-4o, have demonstrated impressive performance in several NLP tasks \cite{cheng2025empowering}. Although these models showed a good performance in English and other high-resource languages, their adaptability to Arabic remains an open question as arabic remains a morphologically rich yet low-resource language \cite{antoun2020arabert}.


This study aims to systematically evaluate reasoning-based LLMs on fifteen critical Arabic NLP tasks. These tasks have been categorized into sentence classification (\textbf{SC}), natural language generation (\textbf{NLG}), and linguistic analysis (\textbf{LA}) tasks. The SC cluster involves sentiment analysis, dialect, hate speech, offensive speech, sarcasm, claim, stance, and paraphrase detection. The NLG cluster involves generation-based tasks, such as summarization, paraphrasing, transliteration, machine translation, and question answering. The LA tasks are word sense disambiguation (WSD) and part-of-speech (PoS) tagging. These tasks have been evaluated under various learning paradigms, such as zero-shot, in-context learning, and fine-tuning. Moreover, we compare the performance of the reasoning-based models with other recent LLMs. Through this comprehensive evaluation, we seek to identify both the potential and the shortcomings of these advanced reasoning-based models in handling Arabic’s linguistic complexities, ultimately guiding further research and development in this area.  


\section{Related Work}

Evaluating the performance of these LLMs requires robust, realistic, and task-specific benchmarks to ensure fair and meaningful comparisons. Due to the unique characteristics of the Arabic language, its complex morphology, rich diacritic system, and numerous dialects, Arabic-specific benchmarks have been developed to better capture these challenges. For instance, diacritization \cite{alkhamissi2020diacretization}, the task of restoring diacritical marks to Arabic text, is a crucial task specific to Arabic. Transliteration \cite{shazal2020transliteration} is another Arabic-specific task, often involving converting Arabizi (romanized Arabic script) into standard Arabic script. Additional tasks and benchmarks tailored to Arabic are discussed in more detail in Section \ref{Sec:tasks}.
Prior studies have examined the performance of LLMs across multiple Arabic NLP tasks. For instance, \cite{elmadany2022orca} introduced recently ORCA, a challenging benchmark for Arabic language understanding, consisting of 60 different datasets across seven natural language understanding tasks. They used ORCA to offer a comprehensive comparison between 16 multilingual and Arabic pretrained language models. However, their evaluation is restricted to BERT-based models, focusing solely on understanding tasks and excluding generative capabilities and newer architectures such as decoder-only LLMs.

Taqyimitw  \cite{alyafeai2023taqyim} assessed the performance of GPT-3.5 and GPT-4 across seven Arabic NLP tasks: sentiment analysis, translation, transliteration, paraphrasing, PoS tagging, summarization, and diacritization, proving the outperformance of GPT-4 on six tasks. They also provide an evaluation pipeline that facilitates the evaluation of such models. However, their study is limited to just two closed-source models from the same family, leaving out a broader range of open-source and multilingual models. Abdelali et al. introduced LAraBench \cite{abdelali2023larabench} for evaluating Arabic NLP and speech processing, comparing multiple models, including GPT-3.5, GPT-4, BLOOMZ, Jais-13b-chat, Whisper, and USM, using zero-shot and few-shot learning techniques to address 33 distinct tasks across 61 publicly available datasets. They found that state-of-the-art (SOTA) models generally outperformed the evaluated models in both zero-shot and few-shot settings, with a few exceptions in few-shot prompting.

Although prior studies such as ORCA, Taqyim, and LAraBench have significantly advanced the evaluation of Arabic NLP, there remains a notable gap in assessing the reasoning capabilities of more recent LLMs. Prior efforts have either focused on NLU using BERT-based models or a limited set of closed-source generative models. In this paper, we address this gap by evaluating a new set of SOTA reasoning-oriented models including DeepSeek-V3 \cite{liu2024deepseek_v3}, DeepSeek-R1 \cite{liu2024deepseek_v3}, GPT-4o, Qwen \cite{bai2023qwen}, and QwQ \cite{qwq32b} on a diverse suite of Arabic NLP tasks.
\section{Benchmark}
\label{sec_benchmark}
\begin{figure*}[tbph]
    \centering
    \includegraphics[width=2\columnwidth]{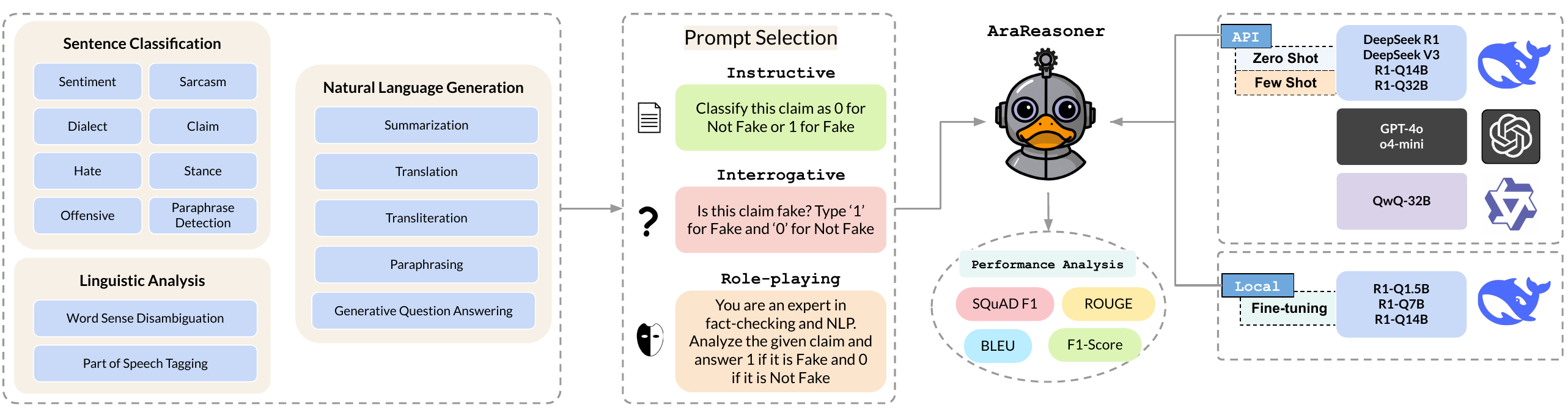}
    \caption{The AraReasoner pipeline.}
    \label{fig:pipeline}
\end{figure*}

Figure \ref{fig:pipeline} shows the pipeline of AraReasoner.
We selected fifteen Arabic NLP tasks to evaluate how well reasoning-based LLMs perform across a diverse range of language understanding challenges. A prompt instruction selection process was followed for each task to identify the most effective prompt (see Appendix \ref{app:prompt_selection}). Subsequently, extensive evaluation experiments were conducted under various prompting strategies, including zero-shot and in-context learning. In addition, we compared the reasoning-based models not only against each other but also against non-reasoning LLMs to highlight the impact of reasoning abilities on task performance.

\subsection{Tasks and Datasets} \label{Sec:tasks}
In this study, we evaluate reasoning-based models on diverse Arabic NLP tasks. Based on their task type, we group these tasks into three categories: SC, NLG, and LA. 
 The dataset statistics for each task are provided in the Appendix (Table \ref{tab:datasets_tasks_single_lines}).


\vspace{1mm}\noindent\textbf{Sentence classification. } This category encompasses a diverse range of sentence-level classification NLP tasks. Eight tasks have been involved in this work: sentiment analysis (\textbf{SA}), dialect detection (\textbf{DD}), sarcasm detection (\textbf{SarD}), hate speech detection (\textbf{HSD}), offensive speech detection (\textbf{OSD}), claim detection (\textbf{CD}), stance detection (\textbf{StD}), and paraphrase detection (\textbf{PD}). 
The SA task involves determining a sentence's emotional tone or opinion, usually as positive, negative, or neutral. We use the AJGT Corpus, which comprises tweets in the Jordanian dialect \cite{alomari2017arabic}, and we used the split proposed by \cite{nagoudi-etal-2022-arat5}. The DD task distinguishes MSA from dialectal Arabic, whereas the SarD task identifies whether a given tweet conveys sarcasm. For both DD and SarD tasks, we utilize the ArSarcasm corpus \cite{farha2020arabic}, which includes annotated labels for both sarcasm and dialect. The HSD is the task of identifying language that expresses hatred, discrimination, or hostility toward individuals or groups based on traits like race, gender, religion, or ethnicity. The OSD task captures rude or disrespectful language that may insult but is not necessarily individual or group-targeted. For HSD and OSD tasks, we employ the dataset proposed by \cite{mubarak2020overview}. The CD task involves checking the factuality of a claim. The StD task detects the writer's stance towards a particular target or topic. The goal is to detect whether the writer agrees, disagrees, discusses, or has no opinion on a given topic.  
For CD and StD tasks, we used the dataset presented in \cite{khouja2020stance} as it provides labels for both tasks. PD checks whether a pair of sentences conveys the same meaning, and we use the dataset introduced by \cite{seelawi2019nsurl} for this task, which consists of Arabic sentence pairs annotated for semantic equivalence.

\vspace{1mm}\noindent\textbf{Natural language generation. } This category includes five principal tasks in \textbf{NLG}. It involves summarization (\textbf{SUM}), paraphrasing (\textbf{PAR}), transliteration (\textbf{TRL}), machine translation (\textbf{MT}), and generative question answering (\textbf{GQA}) tasks. 
The SUM task involves generating a concise and coherent version of a longer text while preserving its key information. For this task, we use the AraSum \cite{kahla-etal-2021-cross} as a training set and EASC as a testing set \cite{el2010using}. This approach was chosen because AraSum provides clear and concise summaries, which enhance fine-tuning quality. The PAR task focuses on rephrasing a given text while maintaining its original meaning. For this task, we utilize the Arabic Paraphrased Parallel Synthetic dataset (APPSD) dataset \cite{al2024arabic}, excluding Arabic Paraphrasing Benchmark (APB) \cite{alian2019towards}, which we use for testing. The TRL task involves converting text from one script to another while preserving its pronunciation. It enables non-Arabic speakers to approximate the pronunciation of Arabic words by representing them using Latin letters. We use the BOLT dataset \cite{shazal2020unified} for this task. We also evaluate the reasoning-based models on the MT task using the Arabic-English split of the United Nations Parallel Corpus v1.0 (UNv1) \cite{ziemski2016united}.  
Moreover, we evaluate these models on the generative QA task that involves generating natural language answers to questions based on a given context. For this task, we utilize the XTREME benchmark \cite{siddhant2020xtreme}.

\vspace{1mm}\noindent\textbf{Linguistic analysis.} This category encompasses POS tagging and  (\textbf{WSD}) tasks. The PoS tagging task (\textbf{PoS}) assigns a grammatical category (e.g., noun, verb, adjective) to each word in a sentence. We use the full Universal Dependencies v2.3 corpus \cite{nivre2018universal} for PoS tagging evaluation experiments. The WSD task identifies the correct meaning of a word with multiple possible senses based on its context. We utilize the Arabic WSD benchmark \cite{el2021arabic} for this task.

\subsection{Model Selection}
This study aims to include a wide range of reasoning-oriented LLMs to comprehensively evaluate their performance on Arabic tasks. Accordingly, we selected a diverse set of models known for their reasoning capabilities, including OpenAI’s o4-mini, DeepSeek R1, and four Qwen-based distilled variants: R1-Q1.5B, R1-Q7B, R1-Q14B, and R1-Q32B. We also included QwQ-32B, a reasoning-focused variant from the Qwen family. For comparison, two non-reasoning models, GPT-4o and DeepSeek v3, were also evaluated, which enables us to contrast reasoning and non-reasoning approaches on the same tasks.

Specifically, we employed GPT-4o, o4-mini, DeepSeek-R1, DeepSeek-V3, R1-Q14B, R1-Q32B, and QwQ-32B for both zero-shot and few-shot prompting to evaluate their out-of-the-box reasoning capabilities under minimal supervision. These models were chosen for their scale and diverse architectural foundations, allowing for a broad comparison across reasoning and non-reasoning paradigms. For fine-tuning experiments, we selected R1-Q1.5B, R1-Q7B, and R1-Q14B, balancing model size and training feasibility. This selection enabled us to explore how different capacities within the same model family respond to task-specific supervision, offering insight into the scalability of fine-tuned reasoning performance.

\subsubsection{OpenAI GPT Models}

We include two proprietary models from OpenAI in our evaluation: \textit{GPT-4o} and \textit{OpenAI-o4-mini} (also referred to as GPT-4-mini or o4-mini). These models represent SOTA commercial LLMs and serve as important baselines for evaluating the capabilities of open-source alternatives.

\vspace{0.5mm}\noindent\textbf{GPT-4o } is OpenAI’s flagship model, optimized for performance across a wide spectrum of reasoning, coding, and multimodal tasks. It combines the strengths of GPT-4 with a more efficient architecture, offering faster inference and improved latency. In our experiments, it serves the purpose of providing a point of comparison with non-open-source models.

\vspace{0.5mm}\noindent\textbf{OpenAI-o4-mini } is a smaller, cost-effective reasoning-based variant of GPT-4. Despite its reduced size, it retains strong performance on core reasoning and alignment benchmarks. In our analysis, o4-mini provides a valuable point of comparison for distilled models like DeepSeek-R1-Q32B, which helps to contextualize the effectiveness of scaling and distillation approaches.

\subsubsection{DeepSeek}

We evaluated the reasoning-based models of the DeepSeek \cite{guo2025deepseek}, particularly DeepSeek-R1 and its distilled Qwen-based variants. These models aim to advance reasoning in language models through reinforcement learning (RL) and effective distillation.

\vspace{0.5mm}\noindent\textbf{DeepSeek-R1-Zero} is trained purely via large-scale RL without supervised fine-tuning. This model demonstrated emergent reasoning behaviors such as self-verification and reflection \cite{guo2025deepseek_r1}. However, it struggles with repetition, readability, and language mixing. 

\vspace{0.5mm}\noindent\textbf{DeepSeek-R1} addresses the limitations of DeepSeek-R1-Zero by introducing cold-start supervised fine-tuning before RL. This resulted in improving stability and performance across math, code, and reasoning tasks \cite{guo2025deepseek_r1}.

\vspace{0.5mm}\noindent\textbf{DeepSeek-V3} DeepSeek-V3 is a 671B-parameter mixture-of-experts (MoE) model with multi-head latent attention and DeepSeekMoE for efficient training. It uses an auxiliary-loss-free multi-token prediction and load balancing for better performance. Trained on 14.8 trillion tokens with supervised fine-tuning and reinforcement learning, it matches top models with stable training \cite{liu2024deepseek}.

\vspace{0.5mm}\noindent\textbf{Qwen Based Distilled DeepSeek Models} We focus on the distilled variants of DeepSeek-R1 that are based on the Qwen2.5 series, including \textit{1.5B}, \textit{7B}, \textit{14B}, and \textit{32B} models. These models transfer reasoning capabilities from DeepSeek-R1 through a distillation process that preserves core performance characteristics while significantly reducing model size. Among them, \textit{DeepSeek-R1-Q32B} has demonstrated competitive performance, reportedly outperforming OpenAI's o1-mini model on several standard benchmarks \cite{guo2025deepseek_r1}. However, it is important to note that their effectiveness in Arabic NLP tasks remains unexplored.

\subsubsection{Qwen}
\vspace{1mm}\noindent\textbf{QwQ-32B} \cite{qwq32b} is a 32B parameter reasoning model from the Qwen series, developed to tackle complex tasks via advanced RL techniques. This model showed strong reasoning capabilities, achieving competitive results on math, coding, and general benchmarks \cite{qwq32b}. Furthermore, it performed on par with much larger models such as DeepSeek-R1 (671B, 37B active) and OpenAI-o1-mini.

\section{Evaluation}

\subsection{Research Questions}
In this study, we aim to address the following research questions:
\begin{itemize}
\item \textbf{RQ1:} How do reasoning-focused LLMs, particularly the DeepSeek family of models, perform on Arabic NLP tasks compared to established models like GPT-4o?
\item \textbf{RQ2:} What impact do different prompting strategies (zero-shot, 3-shot, and 5-shot) have on model performance across various Arabic NLP tasks?
\item \textbf{RQ3:} How does model performance vary across different categories of Arabic NLP tasks, particularly between classification tasks and generation tasks?
\item \textbf{RQ4:} To what extent can model size and architecture influence performance on Arabic NLP tasks?
\end{itemize}


\subsection{Experimental Setup}

In our experiments, we accessed the LLMs via APIs and conducted all fine-tuning locally. For fine-tuning, we focused on select models from the DeepSeek family. Appendix~\ref{sec_prompt_selection} lists the prompt strategies followed in this work, and Appendix~\ref{app:details} gives more information on inference settings, fine-tuning configurations, and hardware specifications.

\subsubsection{Data Split}


The datasets were divided into \textit{Train}, \textit{Dev}, and \textit{Test} splits. The \textit{Dev} set was used for prompt selection. \textit{Train} was used for fine-tuning, while \textit{Test} served for final evaluation. See Appendix~\ref{app:data_stats} for more dataset details.

\subsubsection{Evaluation Metrics}

In this study, each task is evaluated using well-established metrics that align with the nature and goals of the task. This ensures a comprehensive and fair assessment of model performance across different linguistic and generative tasks.

\vspace{1mm}\noindent\textbf{Sentence Classification}
tasks are evaluated using the F1-Score, a widely-used metric that balances precision and recall. This is especially appropriate in scenarios where class imbalance is present.

\vspace{1mm}\noindent\textbf{Natural Language Generation}
tasks are evaluated using metrics that measure the fluency, coherence, and semantic fidelity of the generated text.

For the SUM task, we use the ROUGE-L metric, which evaluates the quality of summaries by comparing the longest common subsequence (LCS) between the generated output and reference summaries. ROUGE-L is particularly effective at capturing sentence-level structure and ensuring that the generated summary retains key content and ordering found in the reference, making it suitable for extractive and abstractive summarization.

For PAR, TRL, and MT tasks, we employ the BLEU score. BLEU measures n-gram overlap between the generated output and reference texts, thereby assessing how closely the system output matches human-written references. While it does not account for semantic equivalence, BLEU is robust at scale and provides reliable corpus-level evaluations, making it especially valuable for comparing multiple model outputs over a large test set.

Finally, for the GQA task, we use the SQuAD-style F1-Score. This metric measures the token-level overlap between the predicted answer and the ground truth, combining both precision and recall. It is especially well-suited for extractive and generative QA settings, ensuring that partial correctness and exact matches are both rewarded.

\vspace{1mm}\noindent\textbf{Linguistic Analysis}
tasks, such as PoS tagging and WSD are evaluated using the F1-Score. This metric provides a balanced evaluation of both accuracy and coverage by integrating precision and recall. It is critical for these tasks where success depends not only on correct identification but also on comprehensive and consistent tagging across the entire dataset.

\begin{table}[]
\caption{Performance comparison across different tasks using different prompts. Highlighted cells indicate the best performance within the respective row. \textbf{INS:} Instructive prompt. \textbf{INT:} Interrogative prompt. \textbf{RP: } Role-playing. \textbf{F1 Score} is used for sentence classification and linguistic analysis; \textbf{ROUGE-L} for SUM; \textbf{SQuAD-F1} for GQA; \textbf{BLEU} for MT, TRL, and PAR.}
\label{tab:prompts_selection}
\resizebox{\linewidth}{!}{%
\begin{tabular}{l|ccc|ccc}
\hline
\multicolumn{1}{c|}{\multirow{2}{*}{\textbf{Task}}} & \multicolumn{3}{c|}{\textbf{English}} & \multicolumn{3}{c}{\textbf{Arabic}} \\
 & \textbf{P1 (INS)} & \textbf{P2 (INT)} & \textbf{P3 (RP)} & \textbf{P4 (INS)} & \textbf{P5 (INT)} & \textbf{P6 (RP)} \\
\hline
\multicolumn{7}{c}{\textbf{\textit{Sentence Classification}}} \\ \hline
SA & 59.20 & 35.62 & 90.66 & \textbf{90.98} & 88.97 & 88.65 \\
DD & 48.53 & 51.48 & 48.99 & 50.43 & \textbf{69.63} & 49.50 \\
HSD & 58.73 & 86.28 & 86.99 & \textbf{89.33} & 87.66 & 89.33 \\
OSD & 54.99 & \textbf{85.47} & 56.76 & 83.97 & 57.15 & 84.56 \\
SarD & 65.89 & 46.3 & 47.74 & \textbf{71.17} & 68.16 & 68.16 \\
CD & 27.73 & 37.74 & 37.18 & \textbf{52.87} & 37.01 & 32.56 \\
StD & 43.37 & 29.88 & 48.59 & \textbf{63.62} & 33.45 & 63.44 \\
PD & 58.01 & 56.07 & 58.06 & 56.01 & \textbf{83.49} & 58.42 \\
\hline
\multicolumn{7}{c}{\textbf{\textit{Natural Language Generation}}} \\
\hline
SUM & 17.39 & \textbf{17.41} & 15.77 & 16.24 & 15.76 & 15.60 \\
PAR & \textbf{13.64} & 12.37 & 8.176 & 10.22 & 9.847 & 4.151 \\
TRL & 0.955 & 1.302 & 0.443 & 1.199 & 1.305 & \textbf{1.445} \\
MT & 5.31 & 3.458 & 5.134 & \textbf{6.03} & 4.813 & 5.663 \\
GQA & 12.4 & \textbf{17.62} & 15.23 & 11.71 & 15.66 & 15.17 \\
\hline
\multicolumn{7}{c}{\textbf{\textit{Linguistic Analysis}}} \\
\hline
WSD & 35.61 & 34.96 & 36.94 & \textbf{58.43} & 38.65 & 57.99 \\
POS & 3.536 & 3.689 & \textbf{3.733} & 3.363 & 3.744 & 3.159 \\
\hline
\end{tabular}}
\end{table}

\subsubsection{Prompts Selection}
We started the evaluation process by selecting the appropriate prompt for each task using the DeepSeek R1-Q14B model. 
We designed six prompt variants for each task by crossing three styles with two languages (English and Arabic) as shown in Appendix~\ref{app:prompt_selection}. The three prompt styles can be classified as \textit{Instructive}, \textit{Interrogative}, and \textit{Role-playing}. Instructive prompts use an imperative tone, Interrogative prompts take the form of a question, and Role-playing prompts present the instruction as if the model is performing a task in a specific role.

Table \ref{tab:prompts_selection} presents the performance of the DeepSeek R1-Q14B model in a zero-shot learning setup, using three prompt types written in both English and Arabic. The results highlight the importance of tailoring prompt styles to each task. As shown in the table, the best performance for most tasks, particularly in SC and LA tasks, was achieved using Arabic instructive prompts. For NLG and some LA tasks, English prompts, especially instructive ones, outperformed the others. Additionally, the lowest performance across nearly all tasks, except for the TRL and PoS tagging tasks, was observed when using role-playing prompts, regardless of language Based on these findings, all subsequent experiments were conducted using the best-performing prompt types(INS, INT, RP) for each task, combined with different few-shot configurations (zero-shot, 3-shot, and 5-shot) to assess the impact of few-shot prompting on task performance.


\subsection{Results}


This section outlines the performance evaluations of the selected LLMs across a variety of Arabic NLP tasks. The findings are detailed in several tables. Firstly, Table \ref{tab:rq1} presents the performance data for 15 tasks, comparing leading reasoning models such as R1-671B, QWQ-32B, and GPT 04-mini, under various prompting strategies. Following this, Table \ref{tab:rq2} illustrates the performance variations across different model sizes within the DeepSeek R1 family for the chosen tasks, also employing diverse prompting strategies. Furthermore, Table \ref{tab:rq3} offers a comparative analysis of the fine-tuning performance of different R1 model sizes on the selected tasks. Finally, Table \ref{tab:rq4} contrasts the performance of non-reasoning models, specifically Deepseek V3-685B and GPT-4o, against the reasoning R1-671B model in a zero-shot setting. These tables collectively provide a detailed view of how model choice, size, and prompting techniques influence outcomes on the benchmarked Arabic NLP tasks. Visual representations of these results can be found in Appendix~\ref{app:graphs}.
 
\begin{table}[t]
\centering
\caption{Reasoning models across tasks. DS: Deepseek; ZS: zero-shot; 3S: 3 shots; 5S: 5 shots.}
\label{tab:rq1}
\resizebox{\linewidth}{!}{%
\begin{tabular}{l|ccc|ccc|ccc}
\hline
\multirow{2}{*}{\textbf{Task}} &
\multicolumn{3}{c|}{\textbf{DS R1-671B}} &
\multicolumn{3}{c|}{\textbf{QwQ-32B}} &
\multicolumn{3}{c}{\textbf{GPT~o4-mini}} \\
& \textbf{ZS} & \textbf{3S} & \textbf{5S} & \textbf{ZS} & \textbf{3S} & \textbf{5S} & \textbf{ZS} & \textbf{3S} & \textbf{5S} \\
\hline
\multicolumn{10}{c}{\textbf{\textit{Sentence Classification}}} \\ \hline
SA   & 88.59 & 87.56 & 87.52 & 35.30 & 85.07 & 87.50 & 60.24 & \textbf{90.18} & 89.61 \\
DD   & 69.42 & 76.76 & 77.79 & 43.49 & 50.20 & 49.01 & 80.69 & 80.50 & \textbf{81.53} \\
HSD  & 63.56 & 62.46 & 64.40 & 39.57 & 39.58 & 61.46 & \textbf{71.14} & 66.69 & 70.11 \\
OSD  & 87.95 & 88.72 & 59.82 & 85.55 & 85.50 & 83.26 & \textbf{90.28} & 88.90 & 87.90 \\
SarD & 59.86 & 65.81 & 67.33 & 22.54 & 36.46 & 38.53 & 66.31 & \textbf{70.96} & 70.32 \\
CD   & 55.43 & 63.11 & 57.72 & 42.46 & 62.75 & \textbf{63.81} & 30.63 & 24.68 & 19.60 \\
StD  & 79.36 & \textbf{95.16} & 88.12 & 72.02 & 87.12 & 82.42 & 76.14 & 85.37 & 75.85 \\
PD   & 88.56 & 89.32 & 91.12 & 56.12 & 56.60 & 86.99 & 91.77 & 91.85 & \textbf{92.45} \\
\hline
AVG  & 74.09 & \textbf{78.61} & 74.23 & 49.63 & 62.91 & 69.12 & 70.90 & 74.89 & 73.42 \\
\hline
\multicolumn{10}{c}{\textbf{\textit{Natural Language Generation}}} \\ \hline
SUM  & 11.70 & 11.79 & 12.07 & 12.37 & 12.93 & \textbf{13.85} & 12.00 & 13.20 & 12.40 \\
PAR  &  4.76 &  8.84 & 10.40 &  7.37 &  9.66 &  9.43 & 14.77 & 16.77 & \textbf{16.80} \\
TRL  & 16.99 & 18.21 & 18.86 &  5.58 &  7.27 &  7.52 & 25.18 & \textbf{28.46} & 28.42 \\
MT   & 24.60 & 26.05 & \textbf{26.39} & 16.86 & 25.24 & 17.01 & 24.02 & 25.03 & 25.00 \\
GQA  & 19.03 & 26.60 & \textbf{29.83} & 12.76 & 17.01 & 19.30 & 20.65 & 23.98 & 27.77 \\
\hline
AVG  & 15.42 & 18.30 & 19.51 & 10.99 & 14.42 & 13.42 & 19.32 & 21.49 & \textbf{22.08} \\
\hline
\multicolumn{10}{c}{\textbf{\textit{Linguistic Analysis}}} \\ \hline
WSD  & 70.30 & 71.43 & \textbf{72.26} & 43.84 & 66.69 & 44.71 & 69.83 & 69.71 & 70.76 \\
POS  &  5.27 & 18.41 & \textbf{24.07} &  3.36 &  8.81 & 11.33 &  6.12 &  7.64 & 10.63 \\
\hline
AVG  & 37.79 & \textbf{44.92} & 48.17 & 23.60 & 37.75 & 28.02 & 37.98 & 38.68 & 40.70 \\
\hline
\end{tabular}%
}
\end{table}

\begin{table}[t]
\centering
\caption{Performance across different model sizes.}
\label{tab:rq2}
\setlength{\arrayrulewidth}{0.5pt}
\setlength{\tabcolsep}{4pt}
\resizebox{\linewidth}{!}{%
\begin{tabular}{l|ccc|ccc|ccc}
\hline
\multirow{2}{*}{\textbf{Task}} &
\multicolumn{3}{c|}{\textbf{R1-Q14B}} &
\multicolumn{3}{c|}{\textbf{R1-Q32B}} &
\multicolumn{3}{c}{\textbf{DS R1-671B}} \\
& \textbf{ZS} & \textbf{3S} & \textbf{5S} &
  \textbf{ZS} & \textbf{3S} & \textbf{5S} &
  \textbf{ZS} & \textbf{3S} & \textbf{5S} \\
\hline
\multicolumn{10}{c}{\textbf{\textit{Sentence Classification}}} \\
\hline
SA   & 82.32 & 86.93 & 87.54 & 86.04 & 87.38 & 84.91 & \textbf{88.59} & 87.56 & 87.52 \\
DD   & 33.16 & 70.22 & 70.87 & 33.68 & 48.50 & 48.98 & 69.42 & 76.76 & \textbf{77.79} \\
HSD  & 40.45 & 59.15 & 61.49 & 41.02 & 62.20 & 62.23 & 63.56 & 62.46 & \textbf{64.40} \\
OSD  & 57.11 & 55.67 & 81.95 & 85.95 & 85.05 & 56.31 & 87.95 & \textbf{88.72} & 59.82 \\
SarD & \textbf{69.74} & 69.12 & 46.88 & 42.14 & 42.17 & 64.96 & 59.86 & 65.81 & 67.33 \\
CD   & 27.43 & 59.20 & 56.11 & 39.97 & 60.23 & 41.19 & 55.43 & \textbf{63.11} & 57.72 \\
StD  & 24.68 & 67.58 & 70.45 & 71.50 & 79.99 & 75.06 & 79.36 & \textbf{95.16} & 88.12 \\
PD   & 41.55 & 52.43 & 87.51 & 59.09 & 87.31 & 88.43 & 88.56 & 89.32 & \textbf{91.12} \\
\hline
AVG  & 47.05 & 65.03 & 70.35 & 57.42 & 69.10 & 65.25 & 74.09 & \textbf{78.61} & 74.22 \\
\hline
\multicolumn{10}{c}{\textbf{\textit{Natural Language Generation}}} \\
\hline
SUM  &  9.43 & 11.37 & 11.98 & 10.57 & \textbf{12.43} & 12.01 & 11.70 & 11.79 & 12.07 \\
PAR  &  7.45 & \textbf{11.04} &  9.18 &  8.38 & 10.80 & 10.30 &  4.76 &  8.84 & 10.40 \\
TRL  &  1.61 &  1.86 &  2.07 &  3.12 &  4.08 &  4.41 & 16.99 & 18.21 & \textbf{18.86} \\
MT   &  9.62 &  9.93 &  9.68 & 13.04 & 15.92 & 13.41 & 24.60 & 26.05 & \textbf{26.39} \\
GQA  & 10.29 & 14.31 & 15.14 & 11.94 & 14.45 & 15.92 & 19.03 & 26.60 & \textbf{29.83} \\
\hline
AVG  &  7.68 &  9.70 &  9.61 &  9.41 & 11.53 & 11.21 & 15.41 & 18.29 & \textbf{19.51} \\
\hline
\multicolumn{10}{c}{\textbf{\textit{Linguistic Analysis}}} \\
\hline
WSD  & 40.43 & 43.26 & 43.74 & 63.47 & 43.45 & 66.07 & 70.30 & 71.43 & \textbf{72.26} \\
POS  &  3.40 &  6.10 &  7.60 &  3.90 &  6.00 &  7.64 &  5.27 & 18.41 & \textbf{24.07} \\
\hline
AVG  & 21.91 & 24.68 & 25.67 & 33.68 & 24.725 & 36.85 & 37.78 & 44.92 & \textbf{48.16} \\
\hline
\end{tabular}%
}
\end{table}

\begin{table}[t]
\centering
\tiny
\caption{Comparing finetuned versions of R1-Q models.}
\label{tab:rq3}
\resizebox{0.8\linewidth}{!}{%
\begin{tabular}{l|rrr}
\hline
\textbf{Task} & \textbf{1.5B} & \textbf{7B} & \textbf{14B} \\
\hline
\multicolumn{4}{c}{\textbf{\textit{Sentence Classification}}} \\ \hline
SA   & 71.89 & 86.52 & \textbf{93.87} \\
DD   & 72.88 & 76.28 & \textbf{79.92} \\
HSD  & 53.72 & 64.47 & \textbf{68.75} \\
OSD  & 72.63 & 83.34 & \textbf{90.59} \\
SarD & 54.52 & 65.03 & \textbf{73.60} \\
CD   & 51.35 & 67.47 & \textbf{68.81} \\
StD  & 59.35 & 86.37 & \textbf{98.02} \\
PD   & 92.71 & 95.48 & \textbf{97.21} \\
\hline
AVG  & 66.13 & 78.12 & \textbf{83.85} \\
\hline
\multicolumn{4}{c}{\textbf{\textit{Natural Language Generation}}} \\ \hline
SUM  & 11.16 & 12.48 & \textbf{19.40} \\
PAR  & \textbf{25.85} & 20.51 & 20.48 \\
TRL  & 47.13 & 52.82 & \textbf{57.77} \\
MT   & 10.48 & 13.73 & \textbf{16.97} \\
GQA  &  2.14 &  4.03 & \textbf{13.40} \\
\hline
AVG  & 19.35 & 20.71 & \textbf{25.60} \\
\hline
\multicolumn{4}{c}{\textbf{\textit{Linguistic Analysis}}} \\ \hline
WSD  & 76.25 & 78.15 & \textbf{86.27} \\
POS  & 70.37 & 87.72 & \textbf{95.99} \\
\hline
AVG  & 73.31 & 82.94 & \textbf{91.13} \\
\hline
\end{tabular}%
}
\end{table}

\begin{table}[t]
\centering
\tiny
\caption{Reasoning vs.\ non-reasoning models in zero-shot mode.}
\label{tab:rq4}
\resizebox{0.8\linewidth}{!}{%
\begin{tabular}{l|rrr}
\hline
\textbf{Task} & \textbf{R1-671B} & \textbf{V3-685B} & \textbf{GPT-4o} \\
\hline
\multicolumn{4}{c}{\textbf{\textit{Sentence Classification}}} \\ \hline
SA   & \textbf{88.59} & 87.13 & 86.71 \\
DD   & \textbf{69.42} & 65.80 & 55.75 \\
HSD  & 63.56 & 65.78 & \textbf{66.21} \\
OSD  & \textbf{87.95} & 84.93 & 87.07 \\
SarD & 59.86 & \textbf{70.76} & 64.84 \\
CD   & 55.43 & 38.34 & \textbf{59.98} \\
StD  & 79.36 & \textbf{80.75} & 35.15 \\
PD   & 88.56 & \textbf{89.99} & 89.29 \\
\hline
AVG  & \textbf{74.09} & 72.94 & 68.13 \\
\hline
\multicolumn{4}{c}{\textbf{\textit{Natural Language Generation}}} \\ \hline
SUM  & 11.70 & 12.92 & \textbf{13.60} \\
PAR  &  4.76 &  5.39 & \textbf{13.62} \\
TRL  & 16.99 & 20.94 & \textbf{25.58} \\
MT   & 24.60 & \textbf{25.40} & 25.23 \\
GQA  & 19.03 & 19.89 & \textbf{20.36} \\
\hline
AVG  & 15.42 & 16.91 & \textbf{19.68} \\
\hline
\multicolumn{4}{c}{\textbf{\textit{Linguistic Analysis}}} \\ \hline
WSD  & \textbf{70.30} & 44.17 & 69.05 \\
POS  &  5.27 &  4.74 &  3.85 \\
\hline
AVG  & \textbf{37.79} & 24.46 & 36.45 \\
\hline
\end{tabular}%
}
\end{table}

\subsubsection{Model Performance Across Task Categories}

As outlined in Table \ref{tab:rq1}, GPT o4-mini demonstrated strong capabilities across most Arabic NLP tasks, particularly excelling in SA (90.18\%), DD (81.53\%), HSD (71.14\%), and PD (92.45\%) tasks. However, this model struggled most with CD and StD, achieving a maximum F1-score of only 30.63\% with zero-shot prompting for CD.

For NLG tasks, DeepSeek models showed competitive performance in MT (BLEU score of 26.39 with 5-shot prompting) but were outperformed by GPT-o4-mini on PAR tasks. The ROUGE-L scores for SUM were relatively consistent across prompting strategies, indicating that this capability might be more inherent to the model's pre-training rather than influenced by in-context examples.

\subsubsection{Impact of Prompting Strategies}
Introducing in-context examples as a prompting strategy consistently improves model performance, particularly on classification tasks, though the benefit expands, and in some cases reverses, when too many examples are provided.
For instance, as shown in Table \ref{tab:rq1}, QwQ-32B’s average F1 on sentence classification tasks leaps from 49.6 \% under zero-shot prompting to 62.9 \% with three examples and further to 69.1 \% with five examples. The most dramatic improvements occur on semantically rich tasks: SA rises from 35.3 \% to 87.5 \%, PD from 56.1 \% to 87.0 \%, and CD from 42.5 \% to 63.8 \% as the number of examples increases. These gains underscore how curated examples help the model internalize task intent and disambiguate complex decisions.

NLG tasks exhibit smaller but reliable benefits from few-shot prompting. DeepSeek R1-671B’s SUM performance (ROUGE-L) increases modestly from 11.70 to 12.07, and its MT quality (BLEU) climbs from 24.60 to 26.39 when moving from zero- to five-shot prompts. PAR and GQA follow similar trends, suggesting that while generative models can leverage examples to refine outputs, the magnitude of improvement is constrained by the open-ended nature of these tasks.

In contrast, LA tasks benefit only marginally from additional examples. WSD sees a slight uptick from 70.3 \% to 72.3 \% F1, and PoS tagging, despite jumping from a mere 5.3 \% to 24.1 \% F1 with five-shot prompting, still reflects significant room for growth in capturing Arabic morphological nuance. These results indicate that models struggle to generalize fine-grained linguistic phenomena solely through few-shot prompting.

Crucially, our findings reveal diminishing—and sometimes negative—returns from larger context windows. Certain tasks, such as StD and DD, experience performance declines when examples increase from three to five. We hypothesize that an excess of examples may introduce conflicting patterns or over-specify the task, impeding the model’s flexibility. 

\subsubsection{Model Size and Architecture Effects}
As demonstrated in Table \ref{tab:rq2}, larger model sizes generally correlate with better performance. The largest DeepSeek R1 variant (671B) consistently outperforms smaller versions across most tasks, with an average performance improvement of 25.17 percentage points compared to the 14B variant in the zero-shot setting. However, the relationship between model size and performance is not strictly linear, as evidenced in Table \ref{tab:rq1} by the QwQ-32B model underperforming by 10.62 percentage points compared to the DeepSeek R1-Q32B despite similar parameter counts.

This suggests that architectural design choices and training methodology play significant roles alongside raw parameter count in determining Arabic language capabilities. The Qwen-based DeepSeek R1-32B variant, for instance, showed stronger performance than might be expected based solely on its parameter count, potentially due to architectural advantages.

\subsubsection{Reasoning vs. Non-Reasoning Models}
In table \ref{tab:rq4}, when comparing the reasoning R1-671B model to non-reasoning models including GPT-4o and V3-685B, we observe mixed results across different tasks. GPT-4o demonstrated superior performance on most NLG tasks, while R1-671B outperforms the non-reasoning models and achieves competitive results on sentence classification and linguistic analysis with an exception on SarD task.

These results suggest that while GPT-4o maintains advantages in most generation tasks, the reasoning R1-671B model show particular promise for Arabic classification and linguistic analysis tasks, potentially due to differences in pretraining data composition or architectural design choices optimized for such tasks.
\section{Conclusion}
In this work, we have delivered the first in-depth benchmark of reasoning-oriented LLMs on 15 Arabic NLP datasets, demonstrating how prompt engineering, model architecture, and scale interact to shape performance across classification, generation, and LA tasks. By showing that just three well-chosen examples can yield double-digit F1 improvements, that DeepSeek’s reasoning-focused design outstrips a top GPT o4-mini baseline by 12 points in zero-shot inference, and that LoRA fine-tuning can add up to 8 additional points over sheer scaling, we provide a clear roadmap for maximizing results on morphologically rich, dialectally varied Arabic data. Our comparative analysis also quantifies the remaining 10–15 point gap between reasoning and non-reasoning models, underscoring where future architectures must evolve. Future work will focus on evaluating additional open-source models, exploring methods to enhance performance on specific tasks, and expanding both the range of datasets and the diversity of tasks considered.
\section{Limitations}

Despite the strengths of this study, several limitations should be acknowledged. First, the evaluation was limited to a subset of available datasets and target tasks, which may not fully capture the linguistic richness and variability of the Arabic language, including its diverse dialects, each with unique characteristics and challenges. Second, only a subset of LLMs was considered, potentially limiting the generalizability of the findings across different model architectures and training paradigms. Third, prompt selection was performed using a single LLM, which may introduce bias. Exploring the impact of prompts on other LLMs is important for a comprehensive evaluation of their performance. Finally, the prompting strategy was limited to three types of prompts. Although this was sufficient to establish baseline comparisons, it may not fully leverage the capabilities of each model. More advanced prompt engineering techniques, such as chain-of-thought prompting or opinion-based prompts, could lead to improved performance and deserve further exploration in future work. Furthermore, while we used a standardized prompt format across all models for consistency, we acknowledge that different models may be optimized for different prompting styles; and that prompt optimization per model is a potential direction for future work.

\section*{Acknowledgments}
\noindent The authors would like to acknowledge the support received from the Saudi Data and AI Authority (SDAIA) and King Fahd University of Petroleum and Minerals (KFUPM) under the SDAIA-KFUPM Joint Research Center for Artificial Intelligence Grant JRC-AI-RFP-14.

\bibliography{custom}

@article{hu2022lora,
  title={Lora: Low-rank adaptation of large language models.},
  author={Hu, Edward J and Shen, Yelong and Wallis, Phillip and Allen-Zhu, Zeyuan and Li, Yuanzhi and Wang, Shean and Wang, Lu and Chen, Weizhu and others},
  journal={ICLR},
  volume={1},
  number={2},
  pages={3},
  year={2022}
}

@inproceedings{shazal2020unified,
  title={A unified model for Arabizi detection and transliteration using sequence-to-sequence models},
  author={Shazal, Ali and Usman, Aiza and Habash, Nizar},
  booktitle={Proceedings of the fifth arabic natural language processing workshop},
  pages={167--177},
  year={2020}
}

@inproceedings{khouja2020stance,
  title={Stance Prediction and Claim Verification: An Arabic Perspective},
  author={Khouja, Jude},
  booktitle={Proceedings of the Third Workshop on Fact Extraction and VERification (FEVER)},
  pages={8--17},
  year={2020}
}

@article{radford2018gpt,
  title={Improving language understanding by generative pre-training},
  author={Radford, Alec and Narasimhan, Karthik and Salimans, Tim and Sutskever, Ilya and others},
  year={2018},
  publisher={San Francisco, CA, USA}
}

@article{team2023gemini,
  title={Gemini: a family of highly capable multimodal models},
  author={Team, Gemini and Anil, Rohan and Borgeaud, Sebastian and Alayrac, Jean-Baptiste and Yu, Jiahui and Soricut, Radu and Schalkwyk, Johan and Dai, Andrew M and Hauth, Anja and Millican, Katie and others},
  journal={arXiv preprint arXiv:2312.11805},
  year={2023}
}

@article{bi2024deepseek,
  title={Deepseek llm: Scaling open-source language models with longtermism},
  author={Bi, Xiao and Chen, Deli and Chen, Guanting and Chen, Shanhuang and Dai, Damai and Deng, Chengqi and Ding, Honghui and Dong, Kai and Du, Qiushi and Fu, Zhe and others},
  journal={arXiv preprint arXiv:2401.02954},
  year={2024}
}

@article{touvron2023llama,
  title={Llama: Open and efficient foundation language models},
  author={Touvron, Hugo and Lavril, Thibaut and Izacard, Gautier and Martinet, Xavier and Lachaux, Marie-Anne and Lacroix, Timoth{\'e}e and Rozi{\`e}re, Baptiste and Goyal, Naman and Hambro, Eric and Azhar, Faisal and others},
  journal={arXiv preprint arXiv:2302.13971},
  year={2023}
}

@article{guo2025deepseek_r1,
  title={Deepseek-r1: Incentivizing reasoning capability in llms via reinforcement learning},
  author={Guo, Daya and Yang, Dejian and Zhang, Haowei and Song, Junxiao and Zhang, Ruoyu and Xu, Runxin and Zhu, Qihao and Ma, Shirong and Wang, Peiyi and Bi, Xiao and others},
  journal={arXiv preprint arXiv:2501.12948},
  year={2025}
}

@article{liu2024deepseek_v3,
  title={Deepseek-v3 technical report},
  author={Liu, Aixin and Feng, Bei and Xue, Bing and Wang, Bingxuan and Wu, Bochao and Lu, Chengda and Zhao, Chenggang and Deng, Chengqi and Zhang, Chenyu and Ruan, Chong and others},
  journal={arXiv preprint arXiv:2412.19437},
  year={2024}
}

@inproceedings{nagoudi-etal-2022-arat5,
    title = "{A}ra{T}5: Text-to-Text Transformers for {A}rabic Language Generation",
    author = "Nagoudi, El Moatez Billah  and
      Elmadany, AbdelRahim  and
      Abdul-Mageed, Muhammad",
    editor = "Muresan, Smaranda  and
      Nakov, Preslav  and
      Villavicencio, Aline",
    booktitle = "Proceedings of the 60th Annual Meeting of the Association for Computational Linguistics (Volume 1: Long Papers)",
    month = may,
    year = "2022",
    address = "Dublin, Ireland",
    publisher = "Association for Computational Linguistics",
    url = "https://aclanthology.org/2022.acl-long.47/",
    doi = "10.18653/v1/2022.acl-long.47",
    pages = "628--647",
}

@article{zhu2009active,
  title={Active learning with sampling by uncertainty and density for data annotations},
  author={Zhu, Jingbo and Wang, Huizhen and Tsou, Benjamin K and Ma, Matthew},
  journal={IEEE Transactions on audio, speech, and language processing},
  volume={18},
  number={6},
  pages={1323--1331},
  year={2009},
  publisher={IEEE}
}

@article{nguyen2022measure,
  title={How to measure uncertainty in uncertainty sampling for active learning},
  author={Nguyen, Vu-Linh and Shaker, Mohammad Hossein and H{\"u}llermeier, Eyke},
  journal={Machine Learning},
  volume={111},
  number={1},
  pages={89--122},
  year={2022},
  publisher={Springer}
}

@inproceedings{kahla-etal-2021-cross,
    title = "Cross-lingual Fine-tuning for Abstractive {A}rabic Text Summarization",
    author = "Kahla, Mram  and
      Yang, Zijian Gy{\H{o}}z{\H{o}}  and
      Nov{\'a}k, Attila",
    booktitle = "Proceedings of the International Conference on Recent Advances in Natural Language Processing (RANLP 2021)",
    month = sep,
    year = "2021",
    address = "Held Online",
    publisher = "INCOMA Ltd.",
    url = "https://aclanthology.org/2021.ranlp-1.74",
    pages = "655--663",
}

@article{al2024arabic,
  title={Arabic paraphrased parallel synthetic dataset},
  author={Al-Shameri, Noora and Al-Khalifa, Hend},
  journal={Data in Brief},
  volume={57},
  pages={111004},
  year={2024},
  publisher={Elsevier}
}

@inproceedings{siddhant2020xtreme,
  title={Xtreme: A massively multilingual multi-task benchmark for evaluating cross-lingual generalization},
  author={Siddhant, Aditya and Hu, Junjie and Johnson, Melvin and Firat, Orhan and Ruder, Sebastian},
  booktitle={Proceedings of the International Conference on Machine Learning 2020},
  pages={4411--4421},
  year={2020}
}

@article{el2021arabic,
  title={Arabic gloss WSD using BERT},
  author={El-Razzaz, Mohammed and Fakhr, Mohamed Waleed and Maghraby, Fahima A},
  journal={Applied Sciences},
  volume={11},
  number={6},
  pages={2567},
  year={2021},
  publisher={MDPI}
}

@inproceedings{ziemski2016united,
  title={The united nations parallel corpus v1. 0},
  author={Ziemski, Micha{\l} and Junczys-Dowmunt, Marcin and Pouliquen, Bruno},
  booktitle={Proceedings of the Tenth International Conference on Language Resources and Evaluation (LREC'16)},
  pages={3530--3534},
  year={2016}
}

@inproceedings{alian2019towards,
  title={Towards building Arabic paraphrasing benchmark},
  author={Alian, Marwah and Awajan, Arafat and Al-Hasan, Ahmad and Akuzhia, Raeda},
  booktitle={Proceedings of the Second International Conference on Data Science, E-Learning and Information Systems},
  pages={1--5},
  year={2019}
}

@article{liu2024deepseek,
  title={Deepseek-v3 technical report},
  author={Liu, Aixin and Feng, Bei and Xue, Bing and Wang, Bingxuan and Wu, Bochao and Lu, Chengda and Zhao, Chenggang and Deng, Chengqi and Zhang, Chenyu and Ruan, Chong and others},
  journal={arXiv preprint arXiv:2412.19437},
  year={2024}
}

@inproceedings{antoun2020arabert,
  title={AraBERT: Transformer-based Model for Arabic Language Understanding},
  author={Antoun, Wissam and Baly, Fady and Hajj, Hazem},
  booktitle={Proceedings of the 4th Workshop on Open-Source Arabic Corpora and Processing Tools, with a Shared Task on Offensive Language Detection},
  pages={9--15},
  year={2020}
}

@article{alkhamissi2020diacretization,
  title={Deep diacritization: Efficient hierarchical recurrence for improved Arabic diacritization},
  author={AlKhamissi, Badr and ElNokrashy, Muhammad N and Gabr, Mohamed},
  journal={arXiv preprint arXiv:2011.00538},
  year={2020}
}

@inproceedings{shazal2020transliteration,
  title={A unified model for Arabizi detection and transliteration using sequence-to-sequence models},
  author={Shazal, Ali and Usman, Aiza and Habash, Nizar},
  booktitle={Proceedings of the fifth arabic natural language processing workshop},
  pages={167--177},
  year={2020}
}

@article{elmadany2022orca,
  title={ORCA: A challenging benchmark for Arabic language understanding},
  author={Elmadany, AbdelRahim and Nagoudi, El Moatez Billah and Abdul-Mageed, Muhammad},
  journal={arXiv preprint arXiv:2212.10758},
  year={2022}
}

@article{alyafeai2023taqyim,
  title={Taqyim: Evaluating arabic nlp tasks using chatgpt models},
  author={Alyafeai, Zaid and Alshaibani, Maged S and AlKhamissi, Badr and Luqman, Hamzah and Alareqi, Ebrahim and Fadel, Ali},
  journal={arXiv preprint arXiv:2306.16322},
  year={2023}
}

@article{abdelali2023larabench,
  title={LAraBench: Benchmarking Arabic AI with large language models},
  author={Abdelali, Ahmed and Mubarak, Hamdy and Chowdhury, Shammur Absar and Hasanain, Maram and Mousi, Basel and Boughorbel, Sabri and Kheir, Yassine El and Izham, Daniel and Dalvi, Fahim and Hawasly, Majd and others},
  journal={arXiv preprint arXiv:2305.14982},
  year={2023}
}

@article{bai2023qwen,
  title={Qwen technical report},
  author={Bai, Jinze and Bai, Shuai and Chu, Yunfei and Cui, Zeyu and Dang, Kai and Deng, Xiaodong and Fan, Yang and Ge, Wenbin and Han, Yu and Huang, Fei and others},
  journal={arXiv preprint arXiv:2309.16609},
  year={2023}
}

@inproceedings{alomari2017arabic,
  title={Arabic tweets sentimental analysis using machine learning},
  author={Alomari, Khaled Mohammad and ElSherif, Hatem M and Shaalan, Khaled},
  booktitle={International conference on industrial, engineering and other applications of applied intelligent systems},
  pages={602--610},
  year={2017},
  organization={Springer}
}

@inproceedings{farha2020arabic, title={From arabic sentiment analysis to sarcasm detection: The arsarcasm dataset}, author={Farha, Ibrahim Abu and Magdy, Walid}, booktitle={The 4th Workshop on Open-Source Arabic Corpora and Processing Tools}, pages={32--39}, year={2020}, organization={European Language Resources Association (ELRA)} }

@inproceedings{mubarak2020overview,
  title={Overview of OSACT4 Arabic offensive language detection shared task},
  author={Mubarak, Hamdy and Darwish, Kareem and Magdy, Walid and Elsayed, Tamer and Al-Khalifa, Hend},
  booktitle={Proceedings of the 4th Workshop on open-source arabic corpora and processing tools, with a shared task on offensive language detection},
  pages={48--52},
  year={2020}
}

@misc{qwq32b,
    title = {QwQ-32B: Embracing the Power of Reinforcement Learning},
    url = {https://qwenlm.github.io/blog/qwq-32b/},
    author = {Qwen Qwen-Team},
    month = {March},
    year = {2025}
}

@article{guo2025deepseek,
  title={Deepseek-r1: Incentivizing reasoning capability in llms via reinforcement learning},
  author={Guo, Daya and Yang, Dejian and Zhang, Haowei and Song, Junxiao and Zhang, Ruoyu and Xu, Runxin and Zhu, Qihao and Ma, Shirong and Wang, Peiyi and Bi, Xiao and others},
  journal={arXiv preprint arXiv:2501.12948},
  year={2025}
}

@article{el2010using,
  title={Using mechanical turk to create a corpus of arabic summaries},
  author={El-Haj, Mahmoud and Kruschwitz, Udo and Fox, Chris},
  year={2010},
  publisher={European Language Resources Association}
}

@article{seelawi2019nsurl,
  title={Nsurl-2019 shared task 8: Semantic question similarity in arabic},
  author={Seelawi, Haitham and Mustafa, Ahmad and Al-Bataineh, Hesham and Farhan, Wael and Al-Natsheh, Hussein T},
  journal={arXiv preprint arXiv:1909.09691},
  year={2019}
}

@misc{nivre2018universal,
  author       = {Joakim Nivre and Mitchell Abrams and Željko Agić and Bengt Ahrenberg},
  title        = {Universal Dependencies 2.3},
  year         = {2018},
  howpublished = {LIN-DAT/CLARIN digital library at the Institute of Formal and Applied Linguistics (ÚFAL), Faculty of Mathematics and Physics, Charles University},
  url          = {https://universaldependencies.org/},
}

@inproceedings{alturayeif2022mawqif,
  title={Mawqif: A multi-label Arabic dataset for target-specific stance detection},
  author={Alturayeif, Nora Saleh and Luqman, Hamzah Abdullah and Ahmed, Moataz Aly Kamaleldin},
  booktitle={Proceedings of the Seventh Arabic Natural Language Processing Workshop (WANLP)},
  pages={174--184},
  year={2022}
}

@article{alyafeai2306taqyim,
  title={Taqyim: Evaluating arabic nlp tasks using chatgpt models. arXiv 2023},
  author={Alyafeai, Z and Alshaibani, MS and AlKhamissi, B and Luqman, H and Alareqi, E and Fadel, A},
year={2023},
  journal={arXiv preprint arXiv:2306.16322}
}

@article{chang2024survey,
  title={A survey on evaluation of large language models},
  author={Chang, Yupeng and Wang, Xu and Wang, Jindong and Wu, Yuan and Yang, Linyi and Zhu, Kaijie and Chen, Hao and Yi, Xiaoyuan and Wang, Cunxiang and Wang, Yidong and others},
  journal={ACM transactions on intelligent systems and technology},
  volume={15},
  number={3},
  pages={1--45},
  year={2024},
  publisher={ACM New York, NY}
}

@article{cheng2025empowering,
  title={Empowering llms with logical reasoning: A comprehensive survey},
  author={Cheng, Fengxiang and Li, Haoxuan and Liu, Fenrong and van Rooij, Robert and Zhang, Kun and Lin, Zhouchen},
  journal={arXiv preprint arXiv:2502.15652},
  year={2025}
}

@article{hayou2024lora_plus,
  title={Lora+: Efficient low rank adaptation of large models},
  author={Hayou, Soufiane and Ghosh, Nikhil and Yu, Bin},
  journal={arXiv preprint arXiv:2402.12354},
  year={2024}
}
\appendix

\clearpage

\section{Data statistics}
\label{app:data_stats}
\begin{table*}
\centering
\small
\caption{Summary of Datasets and Tasks. Most datasets do not include an official validation set; in such cases, we randomly sampled 300 examples from the training set to serve as a development set for the prompt selection stage.}
\begin{tabular}{llllll}
\toprule
\textbf{Dataset} & \textbf{Task(s)} & \textbf{Train size} & \textbf{Dev size} & \textbf{Test size} & \textbf{Full size}\\
\midrule
AJGT & SA & 1,440 & - & 360 & 1800\\
ArSarcasm & DD, SarD & 8,437 & - & 2,110 & 10,547\\
OSACT4 & HSD, OSD & 7,000 & - & 1,000  & 8,000\\
ANS-claim & CD & 3,185 & 987 & 456 & 4,628\\
ANS & StD & 2,652 & 756 & 379 &  3,787\\
NSURL-2019 & PD & 11,997 & - & 3,715 & 15,712\\
AraSum, EASC & SUM & 39,683 & - & 152 & 39,835\\
APPSD & PAR & 15,769 & - & 1,010 & 16,779\\
BOLT & TRL & 60,479 & - & 6,630 & 67,109\\
UNv1 & MT & 99,999 & - & 4,000 & 103,999\\
XTREME & QA & 14,805 & - & 921& 15,726\\
Universal Dependencies v2.3 corpus & POS & 6,075 & 909 & 680 & 7,664\\
Arabic WSD Benchmark & WSD & 24,878 & - & 6,220 & 31,098\\
\bottomrule
\end{tabular}
\label{tab:datasets_tasks_single_lines}
\end{table*}

Table~\ref{tab:datasets_tasks_single_lines} provides an overview of the datasets used in our study, including the task type and the number of examples in each data split. A notable characteristic of many of these datasets is the absence of an official development set. Since prompt-based methods require a separate development set for prompt selection and hyperparameter tuning, we addressed this by randomly sampling 300 examples from the training set to serve as a validation set whenever one was not provided. In addition to prompt selection, the training splits were also used for fine-tuning language models when applicable. For datasets with official validation and test splits, we adhered to the provided divisions. For those without, we ensured that the random sampling for the dev set was done without replacement, preserving a clean separation between training, validation, and test data. This procedure ensured consistency across tasks and datasets, enabling a fair and comparable evaluation setup for both prompt-based and fine-tuned approaches.

\section{Prompt Strategies }
\label{sec_prompt_selection}

We used zero-shot and in-context learning setups to evaluate all models in our selection pool, including GPT-4o, DeepSeek R1, Qwen-based distilled variants, and QwQ-32B. This comprehensive approach allowed us to assess both the reasoning capabilities of full-scale models and the adaptability of the distilled variants under different prompt conditions. Moreover, we fine-tuned three distilled variants of DeepSeek R1: R1-Q1.5B, R1-Q7B, and R1-Q14B. These fine-tuned models were selected due to their open availability and because larger models like the full DeepSeek R1 presented challenges in fine-tuning with our available computational resources.  

\vspace{1mm}\noindent\textbf{Zero-shot learning. }
In the zero-shot setting, models were evaluated using the default parameters provided by the API. No task-specific examples or demonstrations were included in the prompt. This setup served as a baseline to assess how well the models could perform without prior exposure or task conditioning.





\vspace{1mm}\noindent\textbf{In-Context Learning. }For in-context learning, we evaluated models using both 3-shot and 5-shot configurations. Default model parameters from the API were used in this setting as well. To select the samples involved in the in-context learning experiments, we first estimated uncertainty scores using AraBERT’s \cite{antoun2020arabert} representations over the training data.  
The rationale behind using uncertain samples was to expose the model to more informative and challenging instances, which are more likely to enhance reasoning capabilities during inference \cite{nguyen2022measure,zhu2009active}. 
From the top 20 most uncertain samples, we manually selected 5 examples. These examples were then used for both 3-shot and 5-shot prompts.
Manual curation ensured that examples exhibiting high uncertainty due to noise or ambiguity were excluded, focusing instead on those that were difficult yet coherent.
From the top 20 most uncertain samples, we manually selected 5 examples. These examples were then used for both 3-shot and 5-shot prompts.

\section{Prompt Selection}
\label{app:prompt_selection}

\vspace{1mm}\noindent\textbf{Prompt Template.}
We standardized our prompt formatting across tasks using a consistent block structure. Each prompt was composed of the following segments in order: a \textit{System Prompt} block, an \textit{Instruction} block, an optional \textit{Example} block (used in few-shot settings), a \textit{Query} block containing the input question or data for the model, and a final \textit{Response} block that the model is expected to complete. These blocks were clearly delimited with '\#\#\#' to facilitate easy parsing during evaluation. For instance, once an LLM generated its output, we could extract the relevant part by splitting on '\#\#\# Question'. Our templates are illustrated in Figure~\ref{fig:promptTemplate}.

\begin{figure*}[tbph]
    \centering
    \includegraphics[width=2.1\columnwidth]{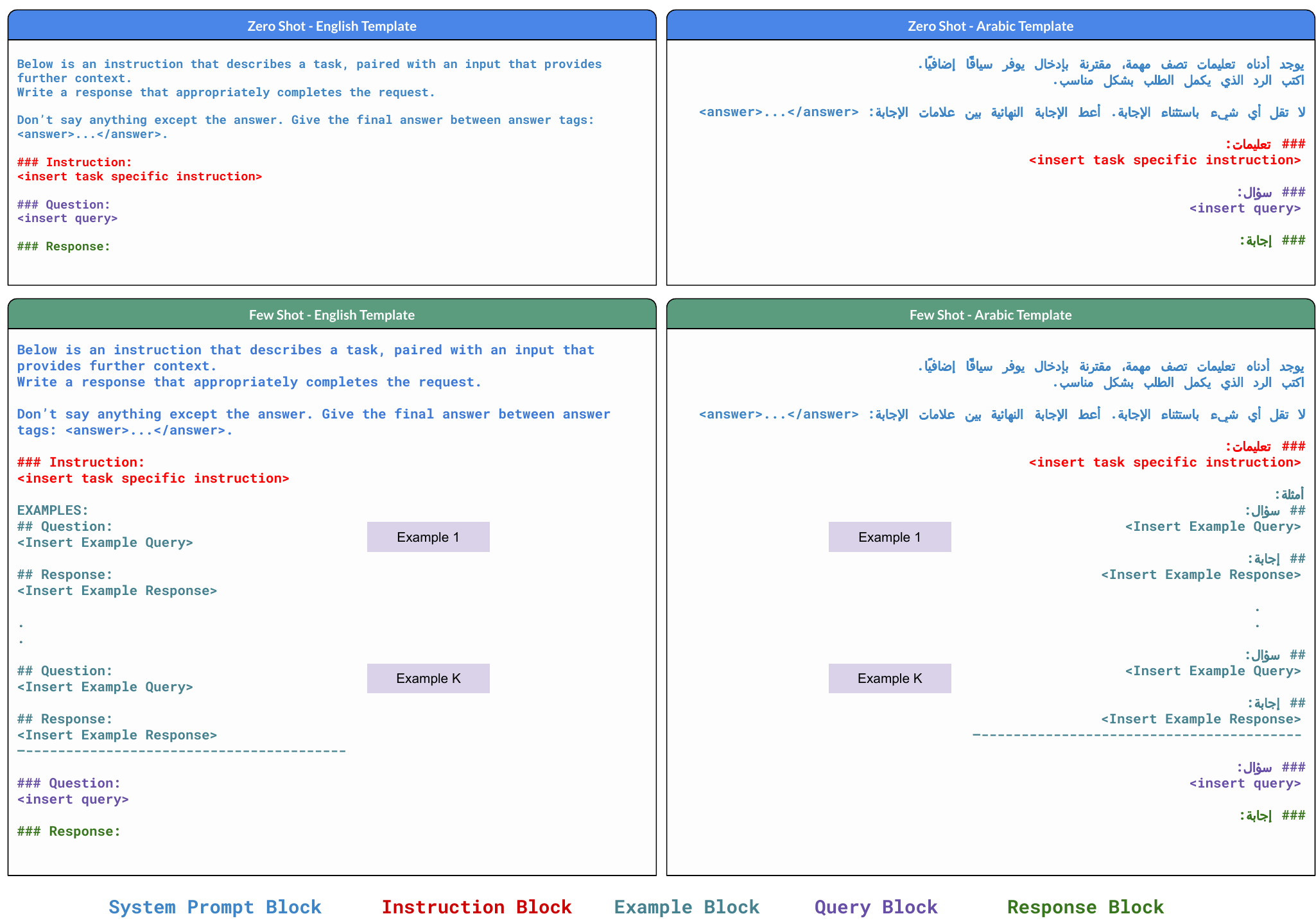}
    \caption{Prompt templates used in finetuning as well as in context learning setup. Zero-shot templates were used in the fine-tuning experiments as well.}
    \label{fig:promptTemplate}
\end{figure*}

\begin{figure*}[tbph]
    \centering
    \includegraphics[width=2.1\columnwidth]{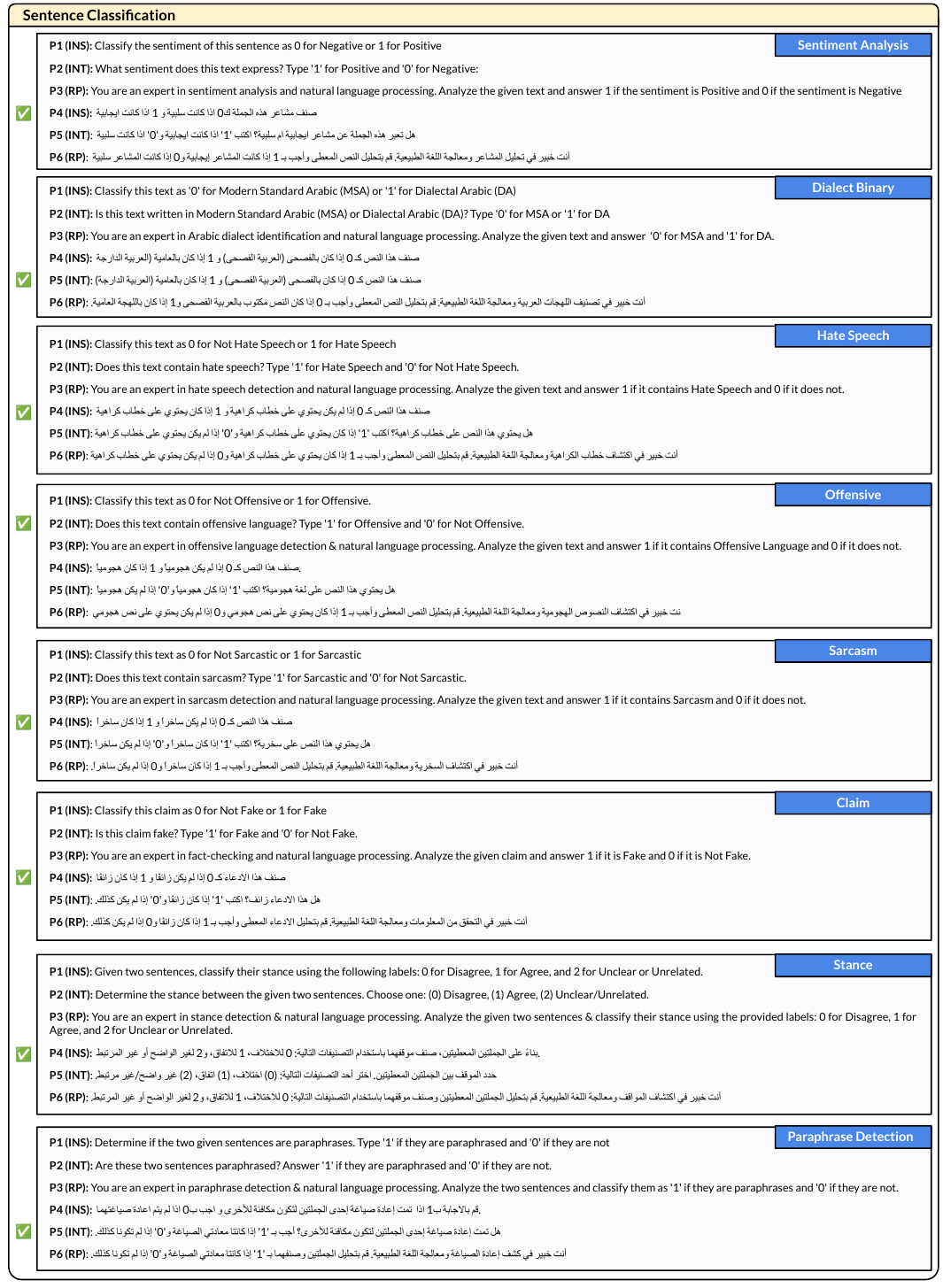}
    \caption{Listing of experimented instructions for sentence classification tasks. INS stands for instructive, INT stands for interrogative, and RP stands for role playing. Ticked instruction indicates the final instruction prompt after instruction comparison.}
    \label{fig:promptSC}
\end{figure*}

\begin{figure*}[tbph]
    \centering
    \includegraphics[width=2.1\columnwidth]{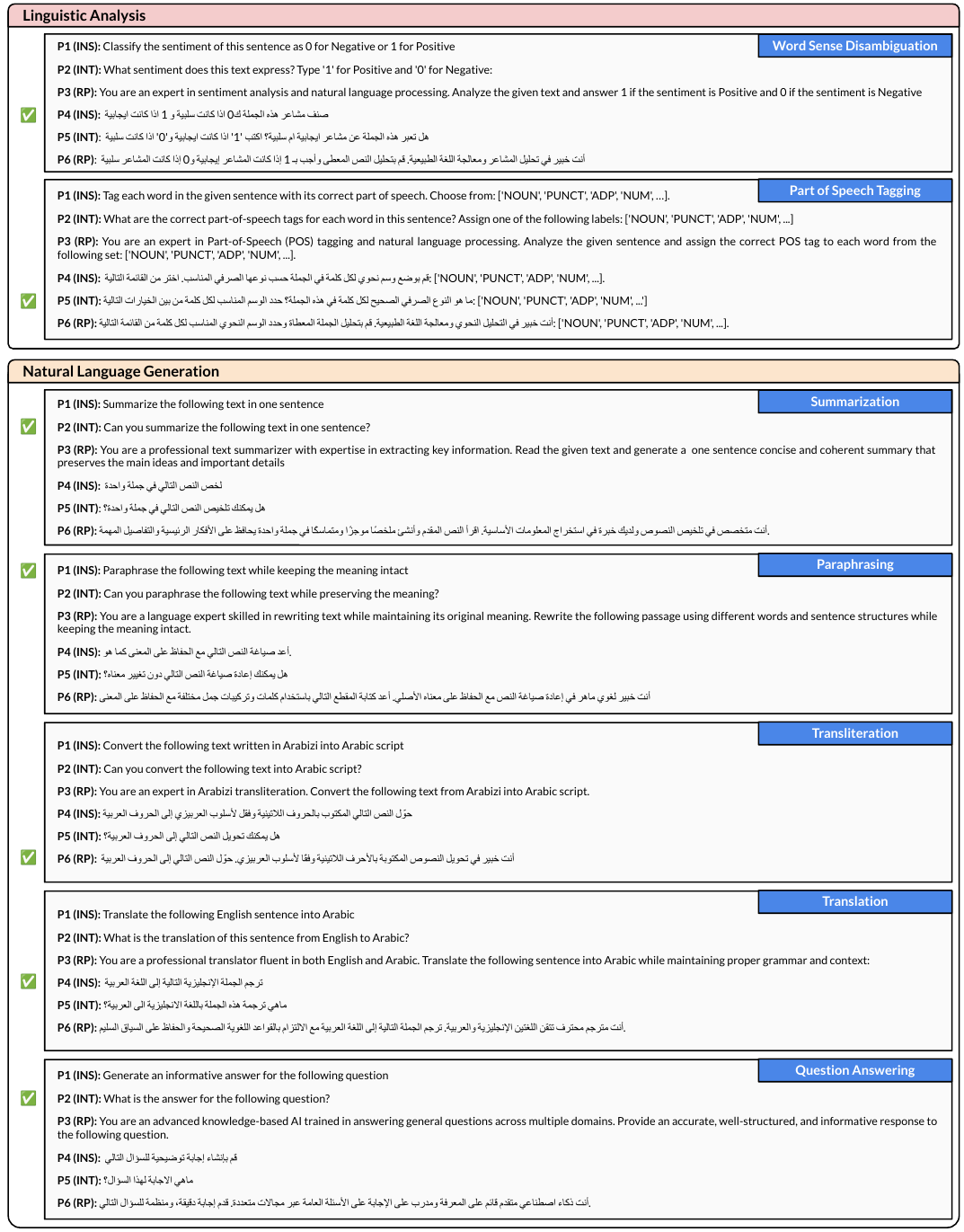}
    \caption{Listing of experimented instructions for Linguistic Analysis and Natural Language Generation tasks. INS stands for instructive, INT stands for interrogative, and RP stands for role playing. Ticked instruction indicates the final instruction prompt after instruction comparison.}
    \label{fig:promptLALG}
\end{figure*}

\vspace{1mm}\noindent\textbf{Instruction Selection.}
For each task, we designed a total of six prompt variants by crossing three styles with two languages (English and Arabic). The three prompt styles were: (1) \textbf{Instructive}, where the prompt uses an imperative tone (e.g., \textit{``Classify the sentiment of the following sentence.''}); (2) \textbf{Interrogative}, where the prompt takes the form of a question (e.g., \textit{``What is the sentiment expressed in this sentence?''}); and (3) \textbf{Role-playing}, which frames the instruction as if the model is performing a task in a role (e.g., \textit{``You are a sentiment analysis expert. Analyze the following sentence and determine its sentiment.''}). Each of these three variants was written in both English and Arabic, resulting in six total prompts per task. To determine the most effective instruction style for each task, we evaluated performance using either the official dev set (if available) or a 300-example subset randomly sampled from the training set (see Appendix~\ref{app:data_stats}). The prompt with the highest score was selected for use in all downstream experiments, including both in-context learning and fine-tuning. See Figure \ref{fig:promptSC} and Figure \ref{fig:promptLALG} for the complete list of the instructions we evaluated.

\vspace{1mm}\noindent\textbf{Instruction Evaluation Results.}
Table~\ref{tab:prompts_selection} summarizes the performance of the six instruction prompts across different tasks. Each row reports the evaluation score for a particular task using each of the six prompt variants: three styles in English (P1–P3) and three in Arabic (P4–P6). As shown, performance varied across both prompt style and language. Arabic prompts frequently outperformed their English counterparts in sentence classification tasks, especially when using an instructive tone (e.g., P4). In general, instructive and interrogative prompts performed better than role-playing prompts in most settings. These findings informed our prompt design decisions in the main experiments.

\vspace{1mm}\noindent\textbf{Selecting Examples for Few-Shot Prompts.}
To select representative and informative examples for few-shot prompting, we adopted an uncertainty-based sampling strategy. Using AraBERT \cite{antoun2020arabert}, we computed uncertainty scores across the training set of each task. Prior work has shown that exposing LLMs to more uncertain and challenging examples can improve performance by encouraging more robust reasoning \cite{nguyen2022measure,zhu2009active}. From the top 20 most uncertain samples, we manually selected 5 high-quality examples. These were used in both 3-shot and 5-shot prompt settings. Manual inspection allowed us to discard noisy or ambiguous examples, ensuring that the selected samples were difficult yet coherent and informative.

\section{Implementation Details}
\label{app:details}

In our experiments, we leveraged two APIs to access the aforementioned LLMs. Specifically, we used the Novita API\footnote{https://novita.ai/} to access both the DeepSeek and QwQ models, and employed the official OpenAI platform\footnote{https://openai.com/api/} for ChatGPT. All models were queried using their default inference settings, except for the maximum token limit, which we set to 2048 tokens. 

For fine-tuning experiments, we focused on a subset of models from the DeepSeek family: R1-Q1.5B, R1-Q7B, and R1-Q14B. We fine-tuned the models using the Low-Rank Adaptation (LoRA) technique \cite{hu2022lora}. In our setup, LoRA was applied to an extended set of projection layers within the Transformer architecture, including the query, key, value, and output projections in the attention mechanism, as well as the gating and feedforward components (specifically, \texttt{q\_proj}, \texttt{k\_proj}, \texttt{v\_proj}, \texttt{o\_proj}, \texttt{gate\_proj}, \texttt{up\_proj}, and \texttt{down\_proj}). This broader application is motivated by findings from \cite{hayou2024lora_plus}, which show that incorporating LoRA into MLP components improves adaptation performance while maintaining parameter efficiency. A LoRA rank of 4 was used. The maximum sequence length was set to 2048 tokens, with a batch size of 2 and gradient accumulation steps of 2, effectively simulating a batch size of 4. Fine-tuning was performed on an NVIDIA RTX A6000 GPU with 48 GB of VRAM. The training ran for up to 100{,}000 steps using the 8-bit AdamW optimizer with a cosine learning rate scheduler and a learning rate of 2e-4. Mixed-precision training was enabled via FP16 or BF16, depending on hardware support.

\section{Evaluation Results}
\label{app:graphs}





\begin{figure*}[h]
\centering
\begin{tabular}{@{}c@{\hspace{0.5cm}}c@{}}
\begin{minipage}{0.465\textwidth}
    \centering
    \includegraphics[width=\linewidth]{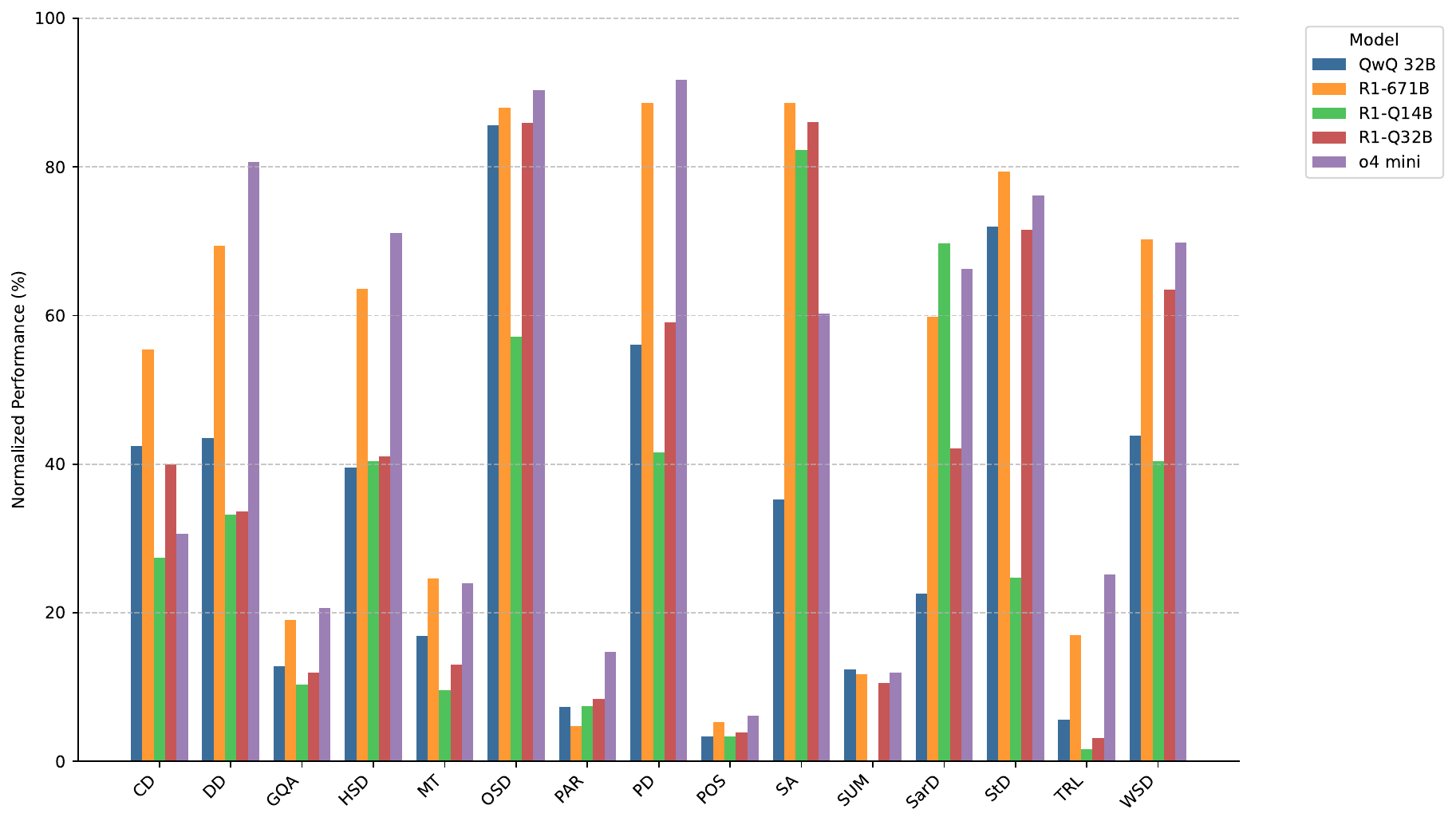}
    \vspace{1mm}
    
    \textbf{(a)}
\end{minipage}
&
\begin{minipage}{0.465\textwidth}
    \centering
    \includegraphics[width=\linewidth]{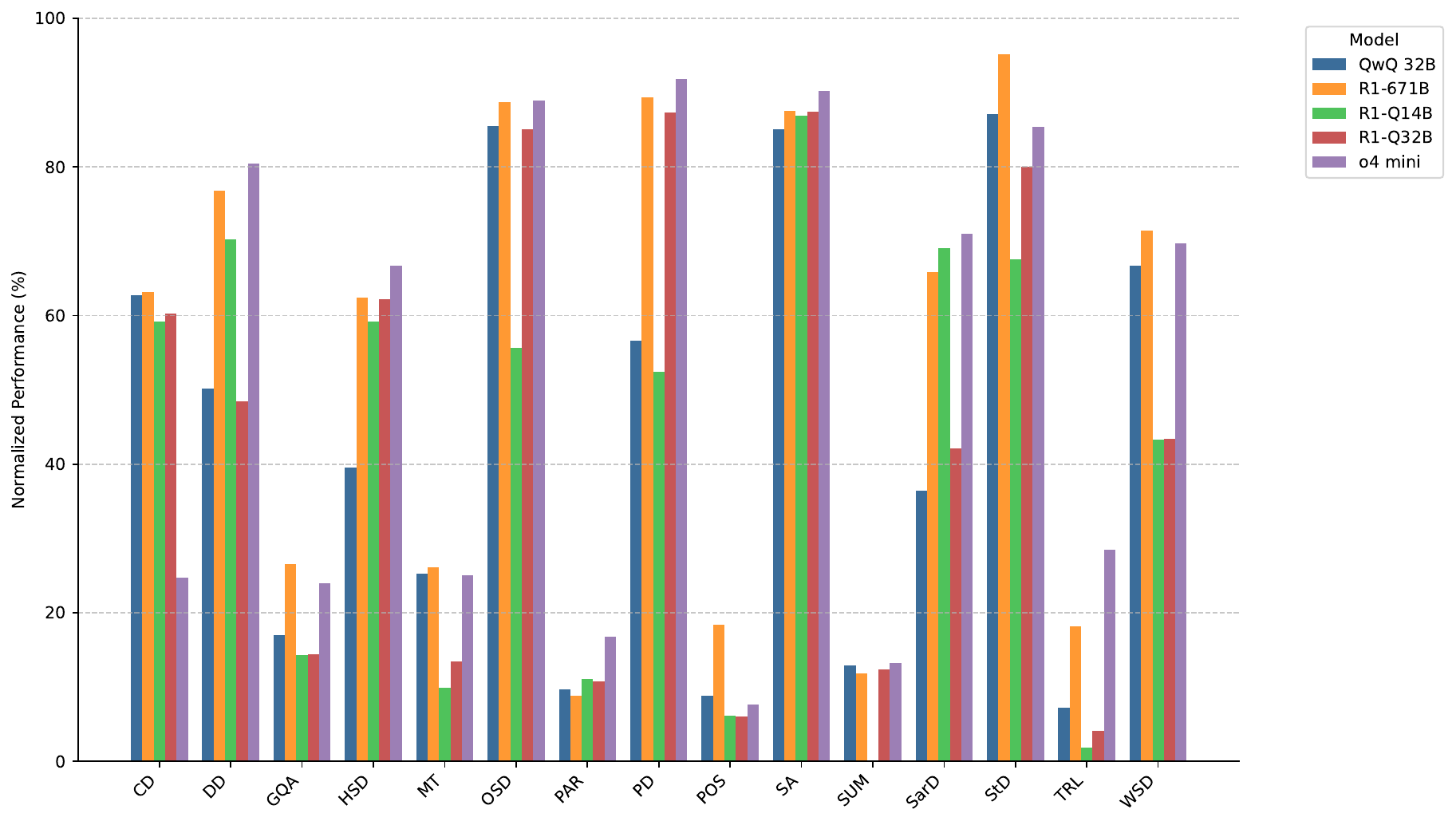}
    \vspace{1mm}
    
    \textbf{(b)}
\end{minipage}
\\[3mm]
\begin{minipage}{0.465\textwidth}
    \centering
    \includegraphics[width=\linewidth]{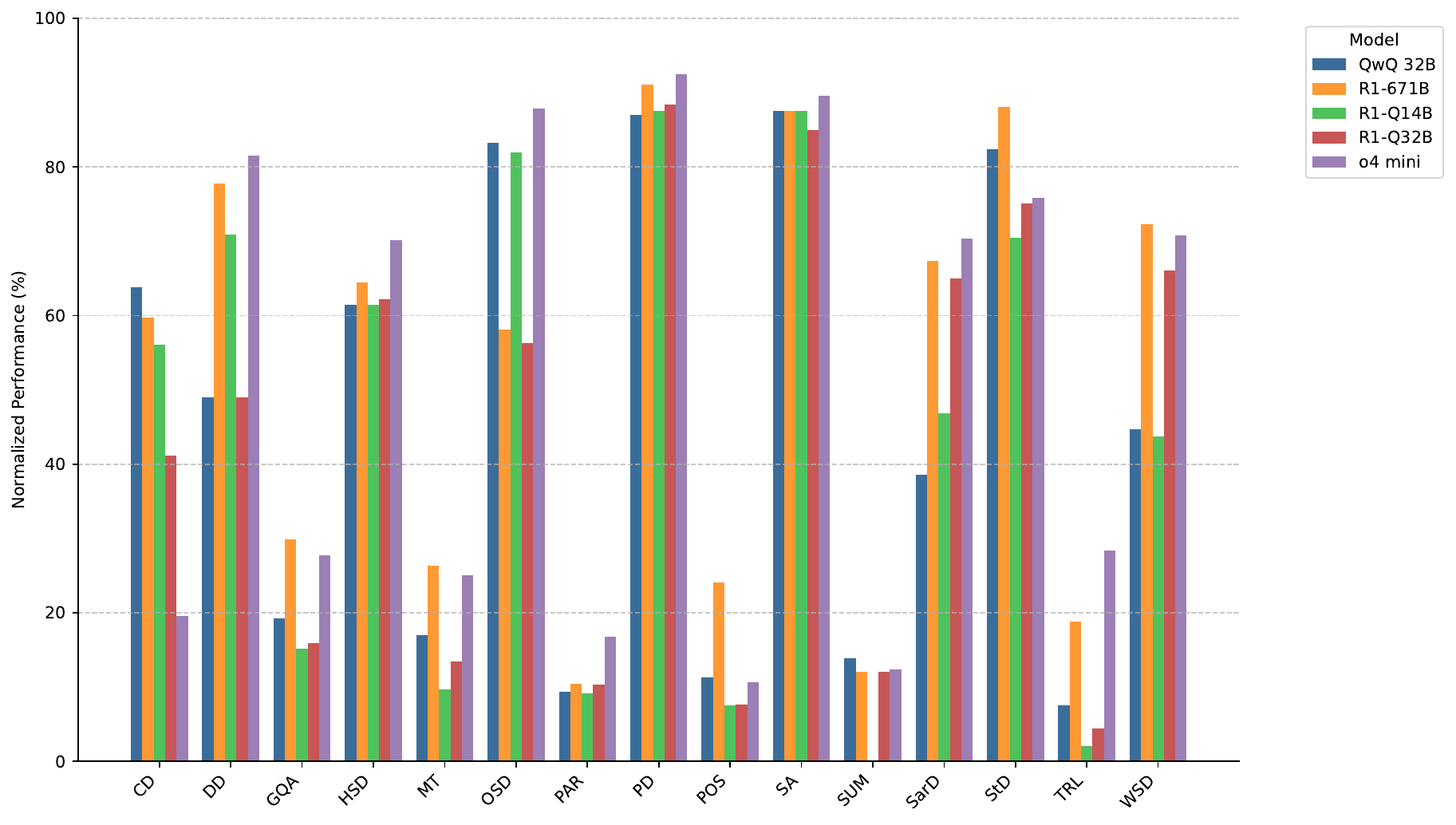}
    \vspace{1mm}
    
    \textbf{(c)}
\end{minipage}
&
\begin{minipage}{0.465\textwidth}
    \centering
    \includegraphics[width=\linewidth]{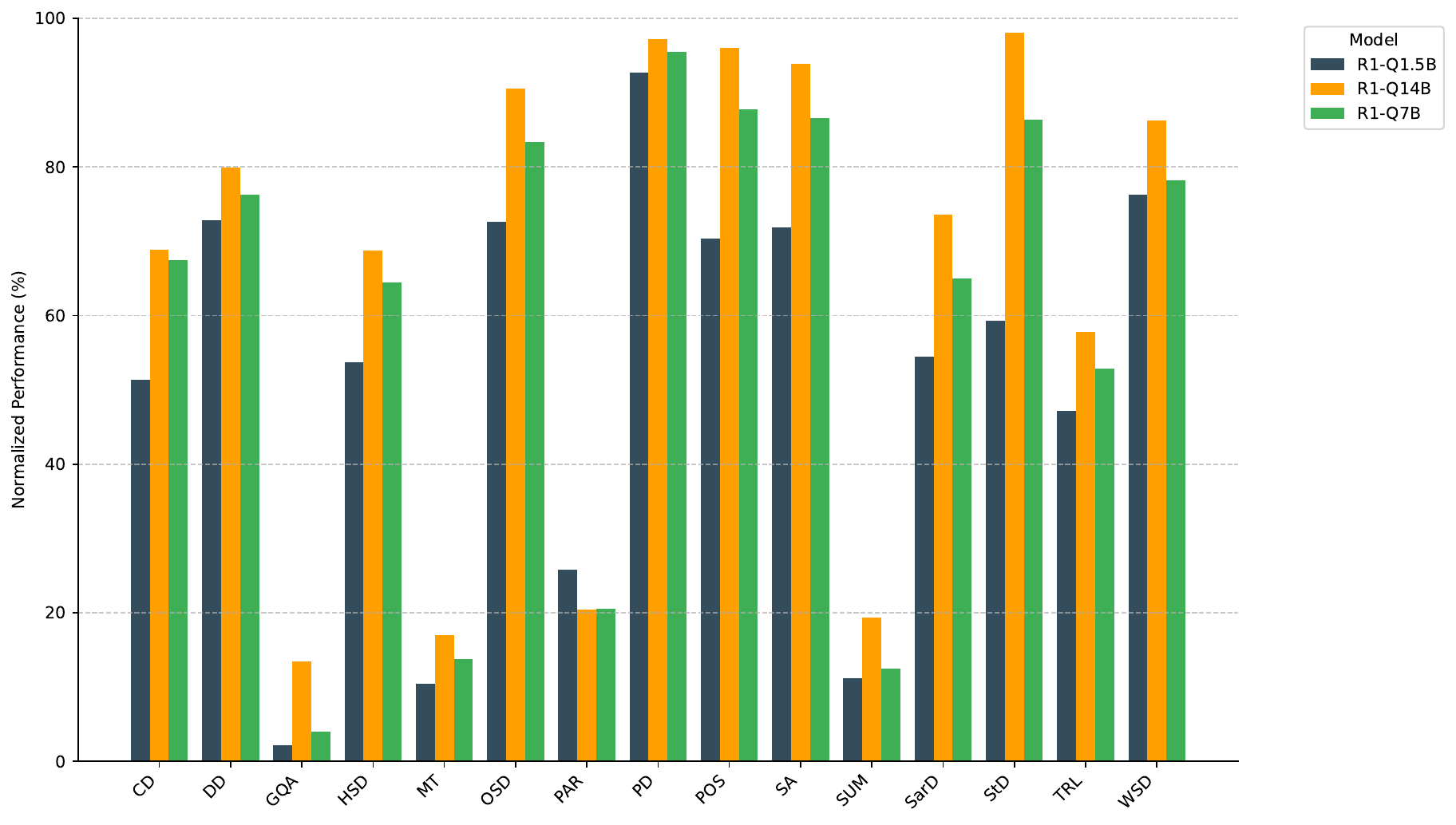}
    \vspace{1mm}
    
    \textbf{(d)}
\end{minipage}
\end{tabular}
\vspace{2mm}
\caption{Normalized performance of the models across different settings: (a) zero-shot, (b) 3-shot, (c) 5-shot, and (d) fine-tuned.}
\label{fig:performance_comparison}
\end{figure*}

Figure~\ref{fig:performance_comparison} compares the normalized performance of the models across four experimental settings: zero-shot, 3-shot, 5-shot, and fine-tuned settings. Figure (a) presents the zero-shot performance when the models are evaluated without being provided with any task-specific examples. The performance varies immensely across tasks, with the largest models generally performing the best due to their better pre-training.

Figure (b) illustrates the 3-shot setting. Providing the model with a few demonstrations improves the performance on the majority of tasks, with greater benefit seen in models with low parameters. This implies that few-shot learning is effective in improving generalization.

Figure (c) presents the 5-shot results. Generally, the trends from the 3-shot remain, with better performance observed in some tasks, particularly those involving classification or structured reasoning. Nonetheless, performance on other tasks degrades relative to 3-shot results.

Figure (d) shows the performance after fine-tuning over a task-specific set. Fine-tuning yields the optimal overall performance for the distilled models, minimizing differences among model sizes and demonstrating the benefit of task adaptation.

\end{document}